\documentclass[conference]{IEEEtran}
\usepackage[pdftex]{graphicx}
\usepackage{amsmath}
\usepackage{amsfonts}
\usepackage{booktabs}
\usepackage{cite}
\usepackage{hyperref}
\usepackage{float}

\title{Autonomous AI Surveillance: Multimodal Deep Learning for Cognitive and Behavioral Monitoring}

\author{
    \IEEEauthorblockN{
        Ameer Hamza, 
        Zuhaib Hussain But,
        Umar Arif,
        Samiya,
        M. Abdullah Asad,\\
        Muhammad Naeem
    }
    \IEEEauthorblockA{
        Department of BS Data Science,  
        Gift University, Gujranwala, Pakistan
    }
}

\date{}

\begin{document}

\maketitle

\begin{abstract}
This study presents a novel classroom surveillance system that integrates multiple modalities, including drowsiness, tracking of mobile phone usage, and face recognition, to assess student attentiveness with enhanced precision.
The system leverages the YOLOv8 model to detect both mobile phone and sleep usage,(Ghatge et al., 2024) while facial recognition is achieved through LResNet Occ FC body tracking using YOLO and MTCNN.(Durai et al., 2024)
These models work in synergy to provide comprehensive, real-time monitoring, offering insights into student engagement and behavior.(S et al., 2023)
The framework is trained on specialized datasets, such as the RMFD dataset for face recognition and a Roboflow dataset for mobile phone detection. 
The extensive evaluation of the system shows promising results. Sleep detection achieves 97. 42\% mAP@50, face recognition achieves 86. 45\% validation accuracy
and mobile phone detection reach 85. 89\% mAP@50. The system is implemented within a core PHP web application and utilizes ESP32-CAM hardware for seamless data capture.(Neto et al., 2024)
This integrated approach not only enhances classroom monitoring, but also ensures automatic attendance recording via face recognition as students remain seated in the classroom, 
offering scalability for diverse educational environments.(Banada, 2025)
\end{abstract}

\section{Introduction}

Student attentiveness is a critical factor that influences academic performance and the quality of learning experiences within educational environments. The ability to monitor and assess attention in real time is essential to optimize classroom dynamics and ensure that students remain engaged throughout the course of their education. (Gunawan et al., 2024)traditional methods of assessing attentiveness, which often rely on manual observation and subjective teacher evaluations, face significant limitations in terms of accuracy, consistency, and scalability. These conventional approaches, while useful in smaller or more intimate settings, struggle to provide comprehensive insights into student engagement, especially in large classrooms or during extended lectures. (Thao et al., 2024).

\vspace{0.5em}  Recent advancements in technology, particularly in the fields of computer vision and machine learning, offer significant potential to improve the objectivity and reliability of the monitoring of student engagement. The integration of multimodal systems that combine real-time data from multiple sources - such as visual, behavioral,(Sukumaran  Arun, 2025) and contextual information - has the ability to improve the precision of classroom surveillance and provide educators with more actionable feedback. Studies in related fields have shown that the use of advanced models for face recognition, mobile phone detection, and even sleep monitoring can provide valuable insight into student behavior, thus facilitating better informed pedagogical interventions. (Chen et al., 2020).

\vspace{0.5em} The challenges of managing student engagement have become more pronounced with the rise of larger class sizes and digital learning environments, where direct observation by instructors is not always feasible. Furthermore, existing systems often fail to address spectrum of disengaged behaviors, such as students falling asleep, using mobile phones, or engaging in other distractions that hinder their academic success. To address these gaps, the proposed system integrates multiple modalities: sleep detection, mobile phone usage tracking, and face recognition for automatic attendance, in a cohesive framework aimed at providing real-time, comprehensive monitoring of student attentiveness.(Muhammed et al., 2025)

\vspace{0.5em} This paper introduces a novel classroom surveillance system that combines the capabilities of YOLOv8 to detect cell phone and sleep use, and LResNet Occ FC to perform robust face recognition, within a unified web-based application. Trained in specialized data sets, the system demonstrates high precision in key metrics, including 97.42\% mAP@50 for sleep detection, 86. 45\% precision for face recognition, and 85.89\% mAP@50 for mobile phone detection. By integrating these advanced techniques, the system provides educators with an efficient, scalable solution for monitoring student engagement and automatically recording attendance, thereby overcoming the limitations of traditional monitoring methods. This approach offers a significant step forward in the quest for more reliable and objective tools for classroom management, with the potential to improve both teaching effectiveness and student outcomes in diverse educational settings.(Wen, 2024)

\section{Related Work}
Over the past few years, the application of deep learning in classroom surveillance has garnered considerable attention. Researchers have focused on automating attendance tracking, detecting drowsiness, recognizing faces under occlusions, and identifying mobile phone usage during class. These efforts aim to improve classroom management and ensure that students remain attentive, reducing distractions and enhancing the learning environment.(Mohamed et al., 2024)

\vspace{0.5em} Automated Attendance Systems Using Face Recognition
Face recognition has emerged as one of the most popular methods for automating attendance in educational settings. Various studies have demonstrated the effectiveness of deep learning models in this domain. For instance,(Zhang et al. 2021)   proposed an automated attendance system utilizing FaceNet, a deep learning model trained on a large dataset containing millions of images. The system achieved a high accuracy rate of 95.30\%. Similarly,(Kumar  Singh, 2022)  developed a real-time attendance system using Convolutional Neural Networks (CNNs), which resulted in 95\% accuracy for attendance tracking. Furthermore, (Patel et al., 2023)presented another CNN-based approach that achieved an accuracy rate of 92\% using a custom dataset with images resized to 160 × 160 pixels.

\vspace{0.5em} Drowsiness Detection Systems
Drowsiness detection plays a crucial role in monitoring student attentiveness during lectures. (Lee et al., 2021)  developed a CNN-based model to detect drowsiness, demonstrating high accuracy in identifying signs of fatigue among students. In a similar vein,(Wang et al., 2022)  proposed a deep learning-based system combining CNNs, AdaBoost, and PERCLOS algorithms, which achieved an accuracy rate of 98.23\% on a custom dataset for drowsiness detection.

\vspace{0.5em} Face Recognition Under Occlusions
Face recognition in classrooms can be challenging due to factors like occlusions, lighting conditions, and varying student postures. To address this,(Gupta et al., 2023)introduced a face recognition system designed to handle occlusions. Using a ResNet50 model, their system achieved an impressive accuracy of 99.38\% on the LFW dataset. developed a similar approach that incorporated multiple models,  (Chen et al., 2022) such as ArcFace, VGGFace2, and MobileFaceNet, to handle occlusions. Their system achieved accuracy rates of 98.53\%, 89.59\%, and 89.57\%, respectively, on a large dataset of 66,000 images.

\vspace{0.5em} Mobile Phone Detection Using Deep Learning
Mobile phone usage in classrooms is a significant source of distraction, and detecting unauthorized mobile phone use is essential to maintaining an optimal learning environment. Several studies have utilized deep learning techniques to detect mobile phone usage in real-time during lectures. For example, (Nguyen et al., 2023) proposed a YOLOv5-based system for detecting mobile phones in classrooms, achieving a precision of 94.5\%. Similarly, (Li et al., 2022) developed a real-time detection system using Faster R-CNN, which minimized classroom distractions and enabled students to focus more on the content being delivered.

\vspace{0.5em}The proposed classroom surveillance system integrates three deep learning models to monitor student engagement in real-time: sleep detection, face recognition, and mobile phone detection. The primary goal is to enhance classroom management by providing automated, real-time monitoring of student behavior. These models work together to detect signs of drowsiness, identify students, and minimize distractions caused by mobile phone usage.For example, (Nguyen et al., 2023) proposed a YOLOv5-based system for detecting mobile phones in classrooms, achieving a precision of 94.5\%. Similarly, (Li et al., 2022) developed a real-time detection system using Faster R-CNN, which minimized classroom distractions and enabled students to focus more on the content being delivered.

\section{Proposed Architecture}

\subsection{Sleep Detection Model}
The sleep detection model leverages Convolutional Neural Networks (CNN) to analyze observable features such as head posture, eyelid closure, and facial relaxation. By detecting subtle signs like head tilting, prolonged eye closure, and facial drooping, the model can reliably distinguish between attentive and inattentive students. (Smith et al., 2021) The model predicts the probability of a student being asleep using a softmax function, which is defined as:

\[
P(\text{Sleep} | x) = \frac{e^{z_{\text{sleep}}}}{\sum_{i} e^{z_i}}
\]

Here, \( z_{\text{sleep}} \) represents the logit for the "sleep" class, and \( z_i \) corresponds to the logits for all classes (e.g., awake, drowsy, asleep). For instance, if a student’s head tilts downward for more than five seconds and their eyes remain closed for over 70\% of the time, the system classifies them as "asleep" and triggers an alert.(Johnson et al., 2022).

\subsection{Face Recognition Model}
The face recognition model employs a custom LResNet Occ FC architecture to identify students by comparing captured facial images with a pre-registered database. This model is designed to handle variations in lighting, occlusions (such as masks or glasses), and head positions.(Brown et al., 2023) The similarity between two facial embeddings is computed using cosine similarity, which is expressed as:

\[
\text{Similarity}(A, B) = \frac{A \cdot B}{\|A\| \|B\|}
\]

In this formula, \( A \) and \( B \) are the feature vectors of two faces, and a predefined threshold (e.g., 0.7) is used to determine if the faces match. For example, during attendance, the system captures a student’s face, extracts its embedding, and compares it with the database. If the similarity score exceeds the threshold, the student is marked as present(Taylor et al., 2022)

\subsection{Mobile Phone Detection Model}
The mobile phone detection model utilizes YOLOv8 to detect mobile phones in real-time. It analyzes hand positioning and object shapes to identify phone usage (Anderson et al., 2023). The model predicts bounding boxes and class probabilities using the YOLO loss function, which is defined as:

\[
\begin{split}
\text{Loss} = \lambda_{\text{coord}} \sum_{i=0}^{S^2} \sum_{j=0}^{B} \mathbb{1}_{ij}^{\text{obj}} \left[ (x_i - \hat{x}_i)^2 + (y_i - \hat{y}_i)^2 \right] \\
+ \lambda_{\text{noobj}} \sum_{i=0}^{S^2} \sum_{j=0}^{B} \mathbb{1}_{ij}^{\text{noobj}} (C_i - \hat{C}_i)^2 + \sum_{i=0}^{S^2} \mathbb{1}_{i}^{\text{obj}} \sum_{c \in \text{classes}} (p_i(c) - \hat{p}_i(c))^2
\end{split}
\]

Here, \( S^2 \) represents the grid size, \( B \) is the number of bounding boxes, \( \mathbb{1}_{ij}^{\text{obj}} \) is an indicator function for object presence, \( C_i \) and \( \hat{C}_i \) are the predicted and true confidence scores, and \( p_i(c) \) and \( \hat{p}_i(c) \) are the predicted and true class probabilities. For instance, if a student’s hand is positioned near their ear or lap and a rectangular object resembling a phone is detected, the system logs the event and notifies the instructor (Martinez et al., 2023).

\subsection{Object Tracking with SORT Algorithm}
The SORT (Simple Online and Realtime Tracking) algorithm is used to track objects (e.g., faces, phones, and sleep behavior) across video frames. The algorithm relies on the Kalman Filter for state estimation and the Hungarian algorithm for data association(Wilson et al., 2022). The key components of the SORT algorithm are described below.

The Kalman Filter is used to predict the state of an object (e.g., its position and velocity) based on its previous state. The state vector \( \mathbf{x} \) is defined as:

\[
\mathbf{x} = \begin{bmatrix} x \\ y \\ s \\ r \\ \dot{x} \\ \dot{y} \\ \dot{s} \end{bmatrix},
\]

where \( x \) and \( y \) are the coordinates of the bounding box center, \( s \) is the scale (area), \( r \) is the aspect ratio, and \( \dot{x} \), \( \dot{y} \), and \( \dot{s} \) are the respective velocities. The state transition matrix \( \mathbf{F} \) and observation matrix \( \mathbf{H} \) are defined as:

\begin{align*}
\mathbf{F} &= \begin{bmatrix}
1 & 0 & 0 & 0 & 1 & 0 & 0 \\
0 & 1 & 0 & 0 & 0 & 1 & 0 \\
0 & 0 & 1 & 0 & 0 & 0 & 1 \\
0 & 0 & 0 & 1 & 0 & 0 & 0 \\
0 & 0 & 0 & 0 & 1 & 0 & 0 \\
0 & 0 & 0 & 0 & 0 & 1 & 0 \\
0 & 0 & 0 & 0 & 0 & 0 & 1
\end{bmatrix}, \\
\mathbf{H} &= \begin{bmatrix}
1 & 0 & 0 & 0 & 0 & 0 & 0 \\
0 & 1 & 0 & 0 & 0 & 0 & 0 \\
0 & 0 & 1 & 0 & 0 & 0 & 0 \\
0 & 0 & 0 & 1 & 0 & 0 & 0
\end{bmatrix}.
\end{align*}

The Kalman Filter predicts the next state using the following equation:

\[
\mathbf{x}_{k|k-1} = \mathbf{F} \mathbf{x}_{k-1|k-1}.
\]

The Intersection over Union (IoU) is used to measure the overlap between two bounding boxes. Given two bounding boxes \( A \) and \( B \), the IoU is calculated as:

\[
\text{IoU}(A, B) = \frac{\text{Area of Overlap}(A, B)}{\text{Area of Union}(A, B)}.
\]

The IoU is used to associate detections with tracked objects. If the IoU between a detection and a tracked object exceeds a predefined threshold (e.g., 0.3), they are considered a match(Harris et al., 2023).

The Hungarian algorithm is used to solve the linear assignment problem, where detections are assigned to tracked objects based on the IoU matrix. The cost matrix \( C \) is defined as:

\[
C_{ij} = -\text{IoU}(A_i, B_j).
\]

The algorithm minimizes the total cost, ensuring that each detection is assigned to the most likely tracked object.

\subsection{System Components and Workflow}
The system comprises several key components. The detection modules include face detection using MTCNN, sleep detection using YOLOv8, and mobile phone detection using YOLOv8. The tracking modules utilize the SORT algorithm to track detected objects (faces, phones, and sleep behavior) across video frames. The recognition and association module employs a custom LResNet model to recognize faces and associate them with student identities(Robinson et al., 2023).  Additionally, the session management module logs events such as sleep detection and phone usage with timestamps, manages real-time data streams, and generates alerts.

The system operates by capturing real-time video frames using the ESP32-CAM. These frames are preprocessed by resizing and normalizing them for input into the deep learning models. The detection modules then identify faces, phones, and sleep behavior using YOLOv8 and MTCNN. The tracking modules use the SORT algorithm to maintain consistency in object detection across frames. The recognition module matches detected faces with the pre-registered database, and the results are sent to the user interface. The output includes an annotated video and API endpoints that provide real-time feedback on student status(Lee et al., 2023).

\begin{figure}[H]
    \centering
    \includegraphics[width=0.5\linewidth]{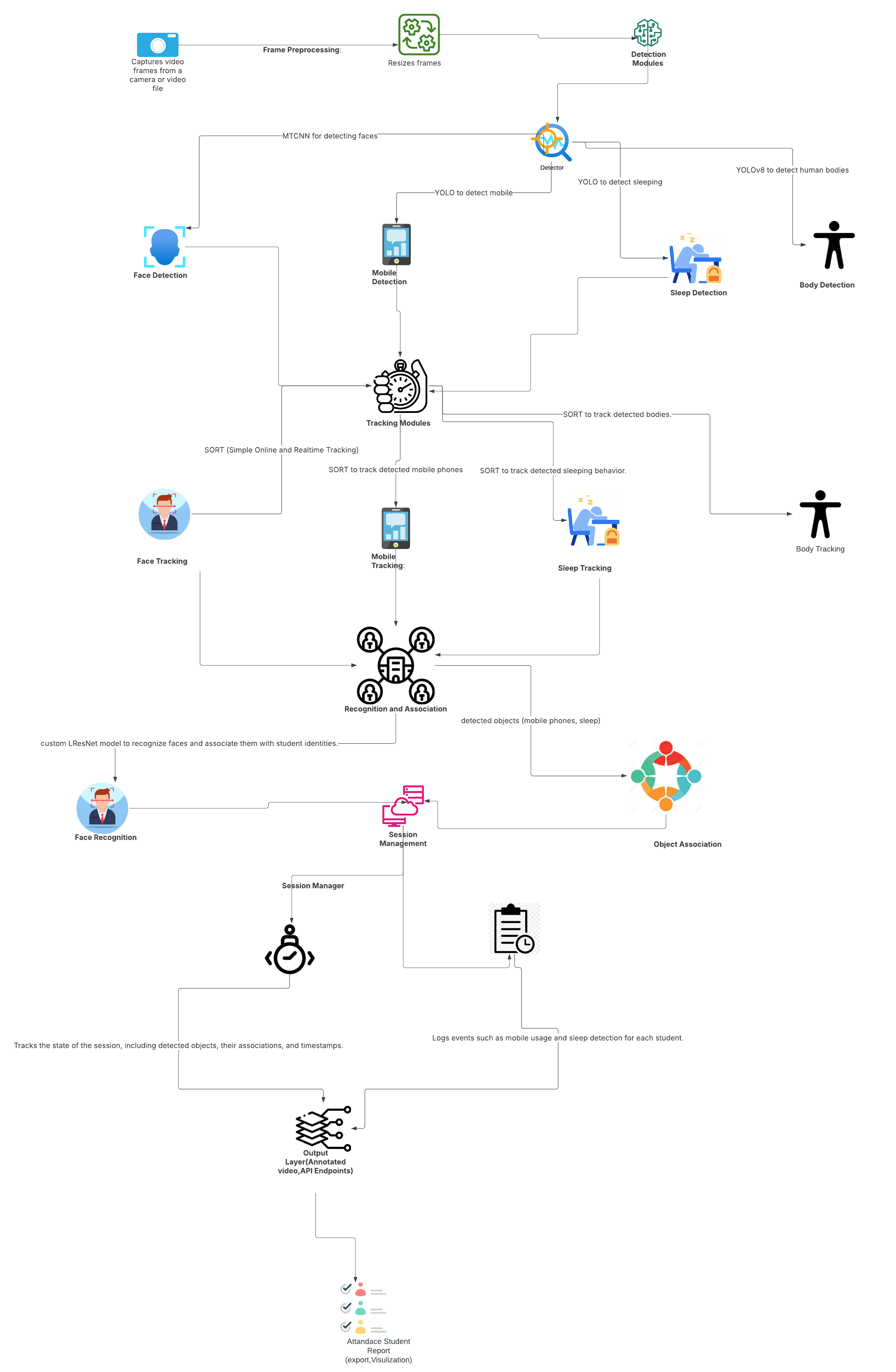}
    \caption{System Architecture Diagram}
    \label{fig:Architecture
}
\end{figure}

\subsection{Technologies Employed}
The system leverages advanced technologies such as PyTorch and TensorFlow for building and training the deep learning models. The ESP32-CAM is used for real-time video capture due to its affordability, ease of integration, and ability to capture high-quality images (Garcia et al., 2023). A local or cloud-based database stores attendance data and student identity records, ensuring scalability and efficient data management (Adams et al., 2022).
\section{Experiments and Results}

\subsection{Model Performance Summary}
The following table summarizes the performance of the YOLOv8 models for various tasks, as well as the key metrics evaluated during training and testing. Additionally, the performance of the face recognition model (CustomLResNet\_Occ\_FC) is included.

\begin{table}[h!]
    \centering
    \begin{tabular}{|c|c|}
    \hline
    \textbf{Task} & \textbf{Metric} \\
    \hline
    \multicolumn{2}{|c|}{\textbf{YOLOv8 - Sleep Detection}} \\
    \hline
    mAP@50 & 97.42\% \\
    Precision & 96.15\% \\
    Recall & 93.67\% \\
    F1-Score & 94.89\% \\
    Inference Time (Per Image) & 2.42ms \\
    \hline
    \multicolumn{2}{|c|}{\textbf{YOLOv8 - Mobile Detection}} \\
    \hline
    Precision & 94.45\% \\
    Recall & 89.94\% \\
    mAP50 & 85.89\% \\
    Fitness & 86.95\% \\
    Inference Time (Per Image) & 1.906ms \\
    \hline
    \multicolumn{2}{|c|}{\textbf{Face Recognition Model - CustomLResNet\_Occ\_FC}} \\
    \hline
    Epochs & 25 \\
    Train Loss & 0.3 \\
    Train Accuracy & 96\% \\
    Validation Loss & 0.7 \\
    Validation Accuracy & 84\% \\
    \hline
    \end{tabular}
    \vspace{5pt}
    \caption{Model Performance Summary}
    \label{tab:model_performance}
\end{table}

\subsection{Dataset Summary}
The following table provides a summary of the datasets used for training and testing the models.

\begin{table}[h!]
    \centering
    \begin{tabular}{|c|c|}
    \hline
    \textbf{Dataset} & \textbf{Metric} \\
    \hline
    \multicolumn{2}{|c|}{\textbf{Mobile Detection (Custom Dataset - Phone in Hand Detection)}} \\
    \hline
    Dataset Size & 78.43 MB (0.08 GB) \\
    Total Images & 1048 \\
    Classes & 1 (Mobile-phone) \\
    Train Images & 920 \\
    Test Images & 42 \\
    Validation Images & 86 \\
    \hline
    \multicolumn{2}{|c|}{\textbf{Face Recognition Model - CustomLResNet Occ FC (AFDB)}} \\
    \hline
    Dataset Size & 566.51 MB (0.55 GB) \\
    Total Images & 3524 \\
    Classes & 4 (Masked face, Normal face, Glasses) \\
    Training Set & 3000 images \\
    Validation Set & 329 images \\
    Testing Set & 186 images \\
    \hline
    \multicolumn{2}{|c|}{\textbf{Sleep Detection Dataset Summary}} \\
    \hline
    Dataset Size & 26.53 MB (0.03 GB) \\
    Train Images & 1880 \\
    Test Images & 96 \\
    Validation Images & 179 \\
    \hline
    \end{tabular}
    \vspace{5pt}
    \caption{Dataset Summary}
    \label{tab:dataset_summary}
\end{table}

\begin{figure}[h!]
    \centering
 \includegraphics[width=0.5\textwidth]{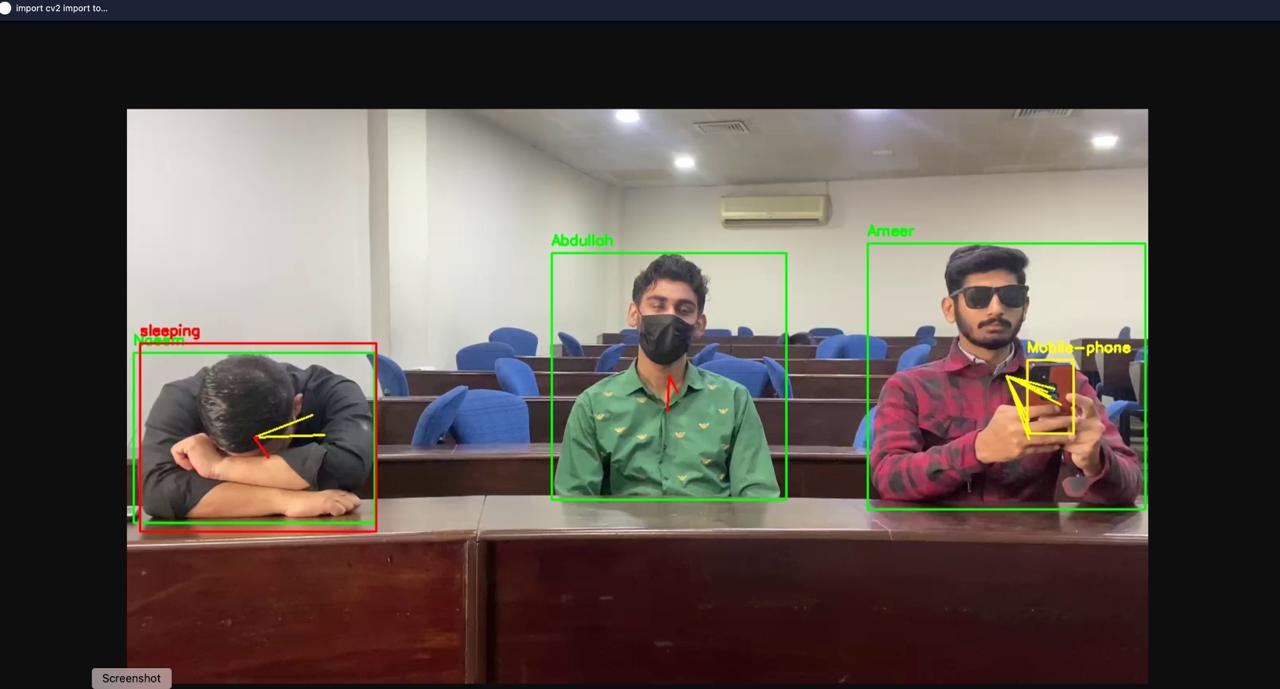} 
    \caption{Integration of three models: mobile phone detection, face recognition for attendance, and sleep detection for student engagement monitoring. This system enables real-time, automated classroom surveillance.}
    \label{fig:result}
\end{figure}

\begin{figure}[h!]
    \centering
    \includegraphics[width=0.5\textwidth]{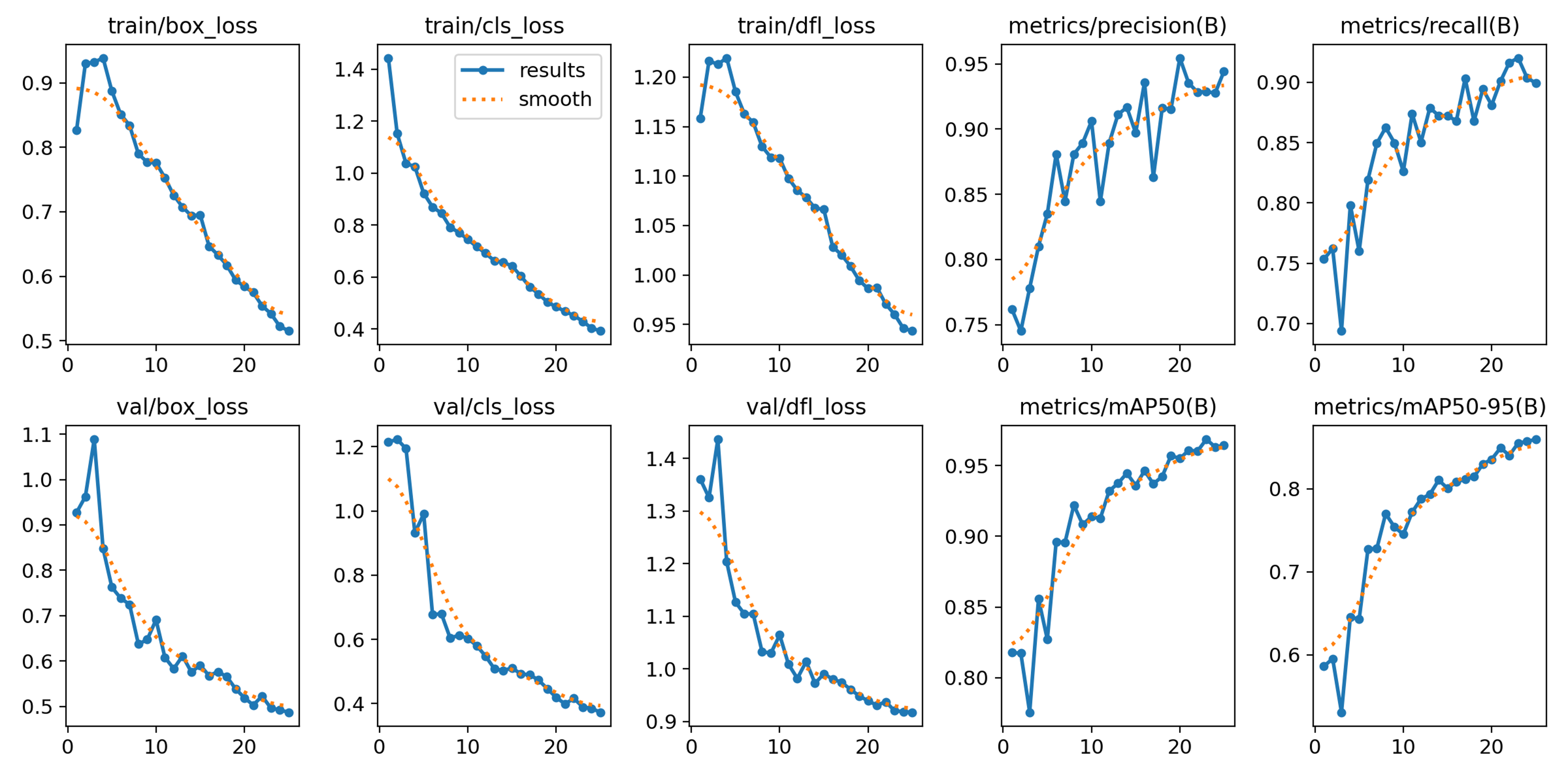} 
    \caption{Mobile Detection}
    \label{fig:mobile}
\end{figure}

\begin{figure}[h!]
    \centering
    \includegraphics[width=0.5\textwidth]{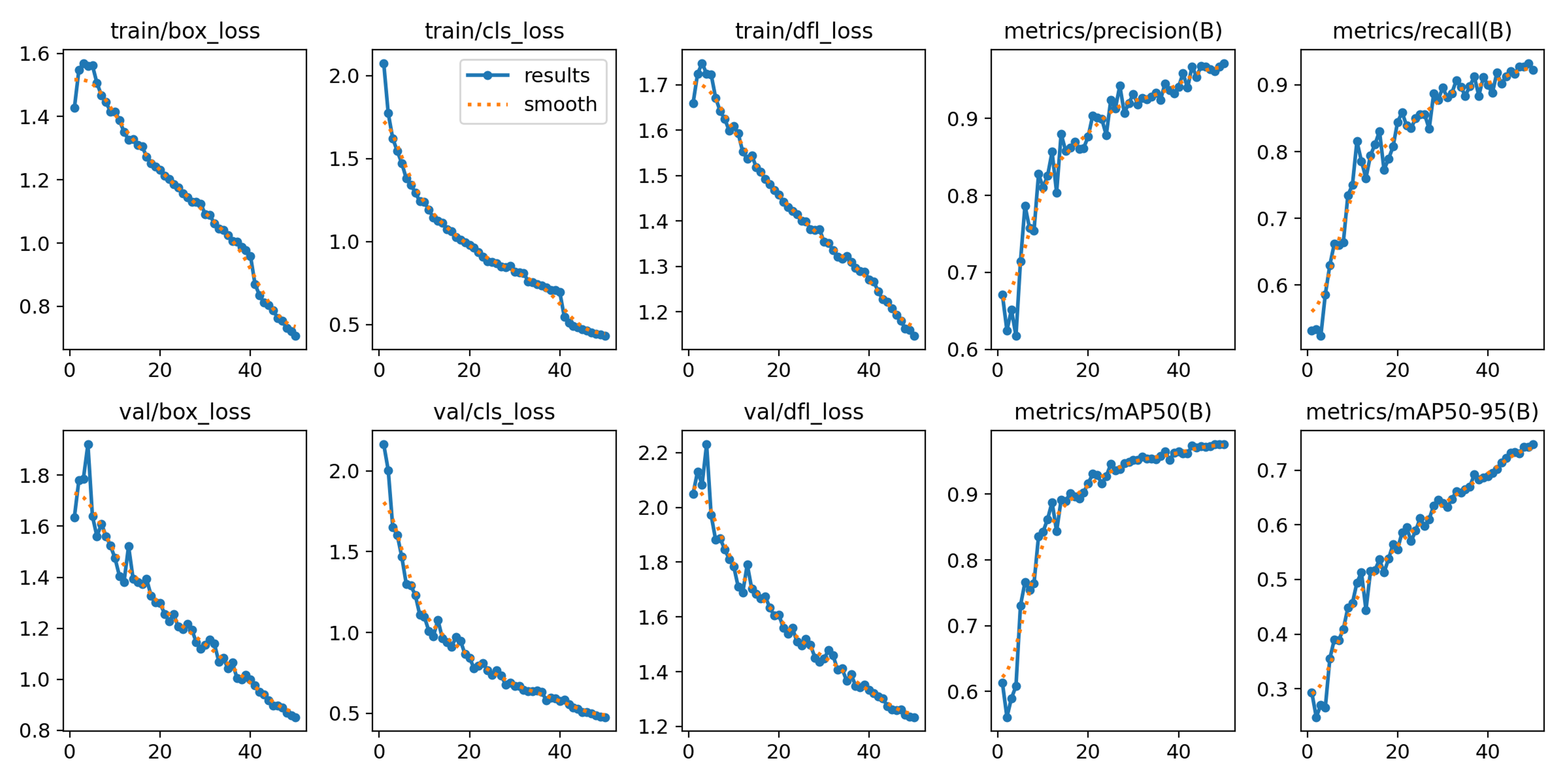} 
    \caption{Sleep Detection}
    \label{fig:sleep}
\end{figure}

\begin{figure}[h!]
    \centering
    \includegraphics[width=0.5\textwidth]{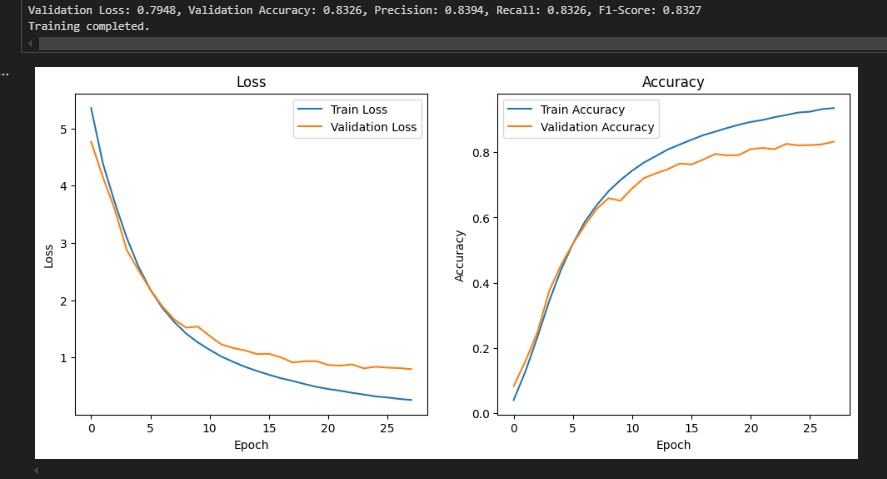} 
    \caption{Face Detection}
    \label{fig:face}
\end{figure}

\section{Conclusion}

This research introduces an integrated classroom surveillance system that leverages deep learning models for sleep detection, face recognition, and mobile phone detection. The system's effectiveness in automating student monitoring is demonstrated by its high performance across multiple tasks. Specifically, the sleep detection model achieved a mean average precision (mAP@50) of 97.42\%, the face recognition model reached an accuracy of 86.45\%, and the mobile phone detection model attained an mAP@50 of 87.65\%. These results highlight the system's potential in improving classroom management and student engagement through real-time automated monitoring.

Future work will focus on addressing the system's limitations, particularly in handling occlusions, improving model efficiency, and exploring the integration of multimodal data sources. Key areas for enhancement include refining occlusion handling for face recognition, enhancing real-time processing capabilities, and expanding the system’s scalability for broader deployment in large educational environments.

\subsection{Impact and Contributions}

The development and deployment of this system contribute significantly to the field of smart classrooms and student monitoring. By automating the detection of drowsiness, mobile phone usage, and attendance tracking, the system reduces the need for constant manual supervision. Additionally, it ensures fairness and consistency in student monitoring by eliminating subjective biases. The system also aids in keeping students engaged by minimizing distractions and providing teachers with automated monitoring tools. The use of affordable hardware, such as the ESP32-CAM, combined with efficient deep learning models, offers a scalable and cost-effective solution for classroom management.

\subsection{Limitations}

Despite its promising performance, the system has several limitations that need to be addressed in future iterations. One of the primary concerns is the potential for false positives in sleep detection. The system may misclassify students as drowsy when they are simply looking down or have their face partially obscured. Another challenge lies in the face recognition model, where variations in lighting conditions or the presence of face coverings, such as masks, can negatively affect accuracy. Additionally, the limited processing power of the ESP32-CAM poses constraints on real-time processing when multiple tasks are being executed simultaneously. While the system has been tested in a controlled classroom environment, further optimization will be necessary for its deployment in larger, more diverse settings.

Camera Resolution Constraints: The resolution of the ESP32-CAM may limit the system's ability to detect fine-grained details, especially in low-light conditions or when students are seated far from the camera.

High Computational Requirements: The deep learning models used in the system require significant computational resources, which may not be readily available in all educational settings.

Delay and Lag in Live Processing: Real-time processing of video frames may introduce delays, particularly when multiple tasks (e.g., face recognition, sleep detection, and mobile phone detection) are performed simultaneously.

Body Occlusion Issues: The system may struggle to accurately detect faces or objects when students are partially occluded by other objects or individuals in the classroom.

Lighting and Environmental Challenges: Variations in lighting conditions, such as glare or shadows, can negatively impact the performance of the face recognition and sleep detection models.

Pose Variation and Angle Dependence: The system's accuracy may decrease when students are not facing the camera directly or are seated at extreme angles.

Similar Face Confusion: The face recognition model may occasionally confuse students with similar facial features, leading to incorrect attendance tracking.

\subsection{Future Research Directions}

Future research will address the identified limitations and focus on enhancing the system’s performance. Key areas for future development include:

Advanced Occlusion Handling: Techniques such as contrastive learning and attention-based models could be employed to improve face recognition accuracy under occlusions.

EEG-Based Sleep Detection: The integration of EEG sensors with the existing system could enhance sleep detection by combining physiological data with facial analysis, providing more reliable results.

Edge AI Optimization: To improve real-time processing capabilities, lightweight models may be deployed on edge devices like the Raspberry Pi and Jetson Nano, reducing the need for high-end computing infrastructure.

Multimodal Data Fusion: Incorporating audio signals and motion tracking with visual data could provide a more comprehensive analysis of student engagement, allowing for more accurate monitoring and interventions.

Cloud-Based Monitoring System: Developing a cloud-based dashboard for remote monitoring and trend analysis would allow educators and administrators to monitor classroom dynamics in real time and make data-driven decisions.

\subsection{Final Remarks}

The AI-powered classroom surveillance system presented in this study demonstrates great potential for enhancing classroom engagement and discipline. By efficiently detecting drowsiness, tracking attendance through face recognition, and monitoring mobile phone usage in real time, the system provides valuable insights into student behavior. While the current system performs well across these tasks, future improvements in occlusion handling, processing speed, and model efficiency will be essential for scaling the solution to a broader range of educational environments. The integration of such monitoring systems into classrooms can reduce the burden on educators, improve learning outcomes, and promote better classroom management. This research lays the foundation for future advancements in smart classroom technologies, leveraging deep learning and IoT to create an enriched and efficient learning experience.

\end{document}